\documentclass[runningheads]{llncs}

\usepackage{cite, makecell}
\usepackage{amsmath,amssymb,amsfonts}
\usepackage{algorithmic}
\usepackage{graphicx}
\graphicspath{{images/}}
\usepackage{textcomp}
\usepackage{xcolor}
\usepackage{balance}
\usepackage{multirow}
\usepackage{hyphenat}

\def\BibTeX{{\rm B\kern-.05em{\sc i\kern-.025em b}\kern-.08em
		T\kern-.1667em\lower.7ex\hbox{E}\kern-.125emX}}

\usepackage{amssymb,amsmath,epsfig,latexsym, bm}
\usepackage{nohyperref}	%pageanchor=false,
\hypersetup{colorlinks, pdfa=true, breaklinks, citecolor=black, urlcolor=black, linkcolor=black}
\usepackage{amssymb,amsmath,epsfig,latexsym, bm}
\usepackage{mathtools, mdwmath, mdwtab, multirow, amssymb, amsmath, array, ragged2e}

\usepackage{threeparttable, booktabs, url, eqparbox, xspace, setspace, tabularx, subfig}
\makeatletter
\DeclareRobustCommand\onedot{\futurelet\@let@token\@onedot}
\def\@onedot{\ifx\@let@token.\else.\null\fi\xspace}
 
\def\ie{\emph{i.e}\onedot}

\def\etal{\emph{et al}\onedot}
\makeatother

\begin{document}
\title{Modelling Lips\hyp State Detection Using CNN for Non-Verbal Communications}
%
% \author{}
 \author{Abtahi Ishmam\and Mahmudul Hasan\and Md. Saif Hassan Onim\orcidID{0000-0002-7228-2823}\and Koushik Roy\and   Md. Akiful Hoque Akif\and Hussain Nyeem\orcidID{0000-0003-4839-5059}}
% %
%\authorrunning{F. Author~\etal}
\authorrunning{A. Ishmam~\etal}
 \institute{Military Institute of Science and Technology (MIST)\\
 Mirpur Cantonment, Dhaka--1216, Bangladesh\\
 \email{abtahiishmam3@gmail.com}\hfill
 \email{mahmud108974@gmail.com}\hfill
 \email{saif@eece.mist.ac.bd}\hfill
 \email{rkoushikroy2@gmail.com}\hfill
 \email{mohammadaxif5717@gmail.com}\hfill
 \email{h.nyeem@eece.mist.ac.bd}\hfill
 }

\maketitle              % typeset the header of the contribution
\begin{abstract}
Vision-based deep learning models can be promising for speech-and-hearing-impaired and secret communications. While such non-verbal communications are primarily investigated with hand-gestures and facial expressions, no research endeavour is tracked so far for the lips state (\ie, open/close)-based interpretation/translation system. In support of this development, this paper reports two new Convolutional Neural Network (CNN) models for lips state detection. Building upon two
prominent lips landmark detectors, DLIB and MediaPipe, we simplify lips-state model with a set of six key landmarks, and use their distances for the lips state classification. Thereby, both the models are developed to count the opening and closing of lips and thus, they can classify a symbol with the total count. Varying frame-rates, lips-movements and face-angles are investigated to determine the effectiveness of the models. Our early experimental results demonstrate that the model with DLIB is relatively slower in terms of an average of 6 frames per second (FPS) and higher average detection accuracy of 95.25\%. In contrast, the model with MediaPipe offers faster landmark detection
capability with an average FPS of 20 and detection accuracy of 94.4\%. Both models thus could effectively interpret the lips state for non-verbal semantics into a natural language.

\end{abstract}
\section{Introduction}
\label{sec-intro}
\begin{figure}
		\includegraphics[width=\columnwidth,height=6cm]{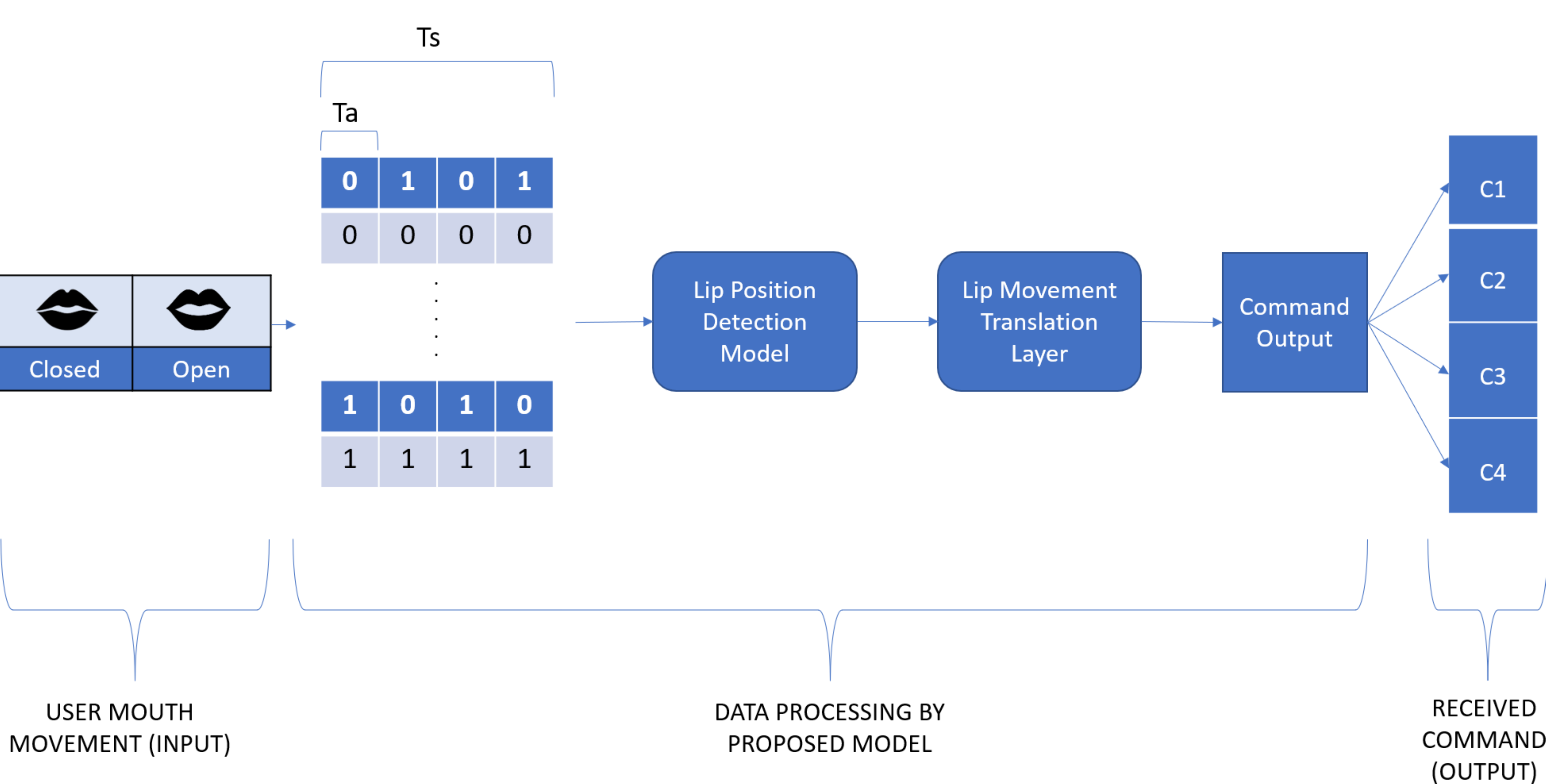}
		\caption{A general framework of the proposed lips\hyp state detection based interpretation/translation model for non\hyp verbal communication system.}
		\label{translation}
	\end{figure}

	Lips\hyp state (\ie, \textit{open} or \textit{close}) detection can be promising for vision\hyp based nonverbal communication system, which has  traditionally been investigated with the head movement and gestures, and facial expression.	Lips\hyp state detection and interpretation is a key step in many security, surveillance and law\hyp enforcement applications.
	For example, lips reading can be helpful in emergency hostage situation. 
	Such communications also enable a speech and hearing impaired person (with a disorder like \textit{stuttering}, \textit{apraxia}, \textit{dysarthria}, or
% 	\textit{speech-sound disorders}, 
    \textit{muteness}) to communicate using lips\hyp state with a minimal effort.
% 	Particularly, several hearing and speech disabilities with either partial or complete impairment to communicate through speech include \textit{stuttering}, \textit{apraxia}, \textit{dysarthria}, 
% % 	\textit{speech-sound disorders}, 
%     \textit{muteness}, \etc. 
% 	In support of the communications of such handicapped people, several Augmentative and Alternative Communication (AAC) systems either aided or unaided have been developed [REF]. 
% % 	\cite{key1, key2, key3}.
% 	
% 	Our research can be categorized as the latter as we have developed a system that can translate simple lip\hyp state combinations into complete instructions resulting in minimization of effort on part of the handicapped user. The system, although simple in application, it has excellent accuracy in instruction translation by simple lips-state detection proving itself reliable in case of application in real-life scenarios with actual human patients. The cost of implementing the proposed system is also undoubtedly negligible as it requires minimal equipment or no equipment at all if integrated into smartphones or tablets. The cost is also low, especially when compared to the massive benefits it may provide to a patient in easing his or her day-to-day activities.
% 
	In support of developing such a system for non-verbal communication  
% 	that would require a minimal effort of the handicapped user and could
	for translating simple lip\hyp state combinations to a complete instruction as illustrated in Fig.~\ref{translation}, we aim to start with the development of a lip\hyp state detection model in this paper.
	The envisaged model with higher possible detection accuracy thus could be promising to a cost-effective and as simple solution as a user-friendly mobile  application.

	Despite having an obvious potential for nonverbal communication, no or a little research endeavours can be tracked in the literature on lips-state  detection. For example, lip-reading was studied for verbal communications using Convolutional Neural Network (CNN) models~\cite{Lu}.
% 	to determine the word spoken from video frames without an audio~\cite{Lu}. 
% 	Lu~\etal~\cite{Lu} presented similar study of lips reading with the help of CNN based feature extraction. With the attention based Long\hyp Short \hyp Term\hyp Memory (LSTM) model their proposed algorithm achieved near ideal result while recognizing mathematical digit pronunciation.
	Besides, lips states and movements have been partially considered for the facial detection and emotion recognition, both with and without the landmarks of lips.
	The case of without lips landmarks
% 	, image frequency based $log$ polar spectrum of the whole image is mainly considered. 
	has higher dependency on the image contrast and spatial resolution, and thus, performance of this approach significantly varies~\cite{bouvier}. 
% 	and remains mostly unreliable~\cite{bouvier}.  
% 	Despite an Support Vector Machine (SVM) based model~\cite{bouvier} is reported to tackle the spatial-domain dependency.
% 	, the overall state detection performance is still not satisfactory.
% 	detection methods without finding the landmarks has also been investigated, where 
% 	the main driver was . This method  Although
% 
	In contrast, the lips landmarks based detection is widely investigated facial detection and emotion recognition.  
% 	Our research, therefore, can be categorized as the latter as we aim to 
% 	The system, although simple in application, it has excellent accuracy in instruction translation by simple lips-state detection proving itself reliable in case of application in real-life scenarios with actual human patients. 
% 	The cost of implementing the proposed system is also undoubtedly negligible as it requires minimal equipment or no equipment at all if integrated into smartphones or tablets. The cost is also low, especially when compared to the massive benefits it may provide to a patient in easing his or her day-to-day activities.
% 	
	For example, Sharma~\etal~\cite{sharma2016farec}  considered face alignment and feature extraction in their face recognition model using Convolutional Neural Network (CNN) and DLIB face alignment.\looseness -1
	
	To better tackle the challenges in lips detection, including varying skin colour, appearance and different lighting conditions, Juan~\etal~\cite{wenjuan2010real} considered the relative distances among faces, eyes and mouth to locate the mouth region. Their model can segment the lips more efficiently than some other prominent models. Amornpan~\etal~\cite{amornpan2019face} applied transfer learning for feature extraction for a face recognition application using a pre-trained deep learning model and validated with the public face datasets like Extended Yale Face Database~B (Cropped) and the Extended Cohn-Kanade Dataset~(CK+).
	
	Similarly, Singh~\etal~\cite{singh2016improved} proposed to use Viola Jones algorithm to localise the face and mouth in the image with an iterative and adaptive construction of merging threshold to make the model image quality invariant. Krause~\etal~\cite{krause2020automatic} introduced a highly portable, and automatic solution for extracting oral posture from digital video using an existing face-tracking utility and OpenFace2. Later, Xu~\etal~\cite{xu2020anchorface} improved the facial landmark localisation across large poses using a split-and-aggregate strategy.
	
	The above vision\hyp based models focus on the face tracking and have a partial consideration of lips detection. In addition to the contextual information extracted from the landmarks of eyes, nose and face-shape, lips were considered to complement the features required for the face detection and recognition and emotion classification. However, as mentioned in the beginning of this section, no lips\hyp state detection is considered so far for developing the envisaged translation model for non\hyp verbal communication system.\looseness -1
	
	In this paper, two new lips\hyp state  detection models have been proposed for non\hyp verbal communications. Both models use landmark detectors for the localisation of lips. Two  highly accurate and robust facial landmark detectors: DLIB~\cite{dlib09} and MediaPipe~\cite{lugaresi2019mediapipe} are considered, which can gather facial landmark information from both real-time and static images. 
	Building upon these detectors, we simplify modelling the lips\hyp state with a set of six key landmarks and their distances to detect relative variations in lips.

Thereby, we develop two models to accurately track the lips\hyp states. Particularly, these models start with the detection of human faces from the captured frames, and isolate the face region followed by the extraction of the landmarks of lips. This identified landmarks and the distance among the landmark points are then used for the successive decision making,  approximation and classification of the lips-state. Performance of these proposed models are finally analysed to learn their merits for the non\hyp verbal communication system (see Sec.~\ref{Result}).

\section{New Lips\hyp State Detection Models}
\begin{figure*}[!thbt]
		\centering
		\subfloat[]{\includegraphics[width=1.1\columnwidth, height=1.35in]{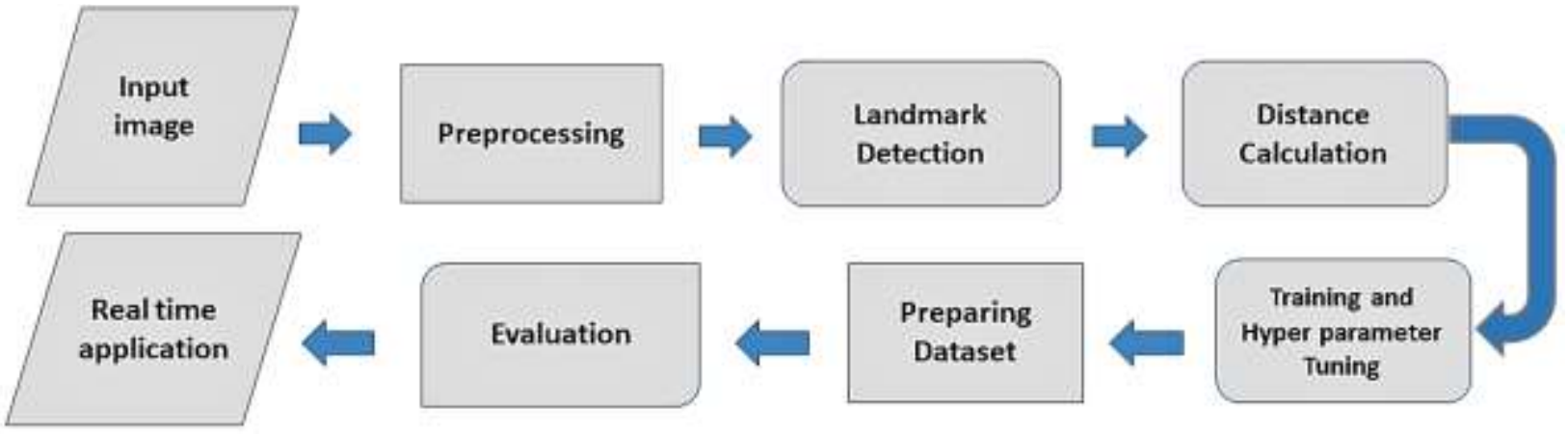}
			\label{detailed flow}}\hfill
		%\caption{Detailed flow diagram for illustrating the methodology.}
		%
		%\end{figure*}
		%\begin{figure}
		%\centering
		% \includegraphics[width=1\linewidth, height=7cm]{model train.png}
		\subfloat[]{\includegraphics[width=0.8\textwidth,height=2.43in]{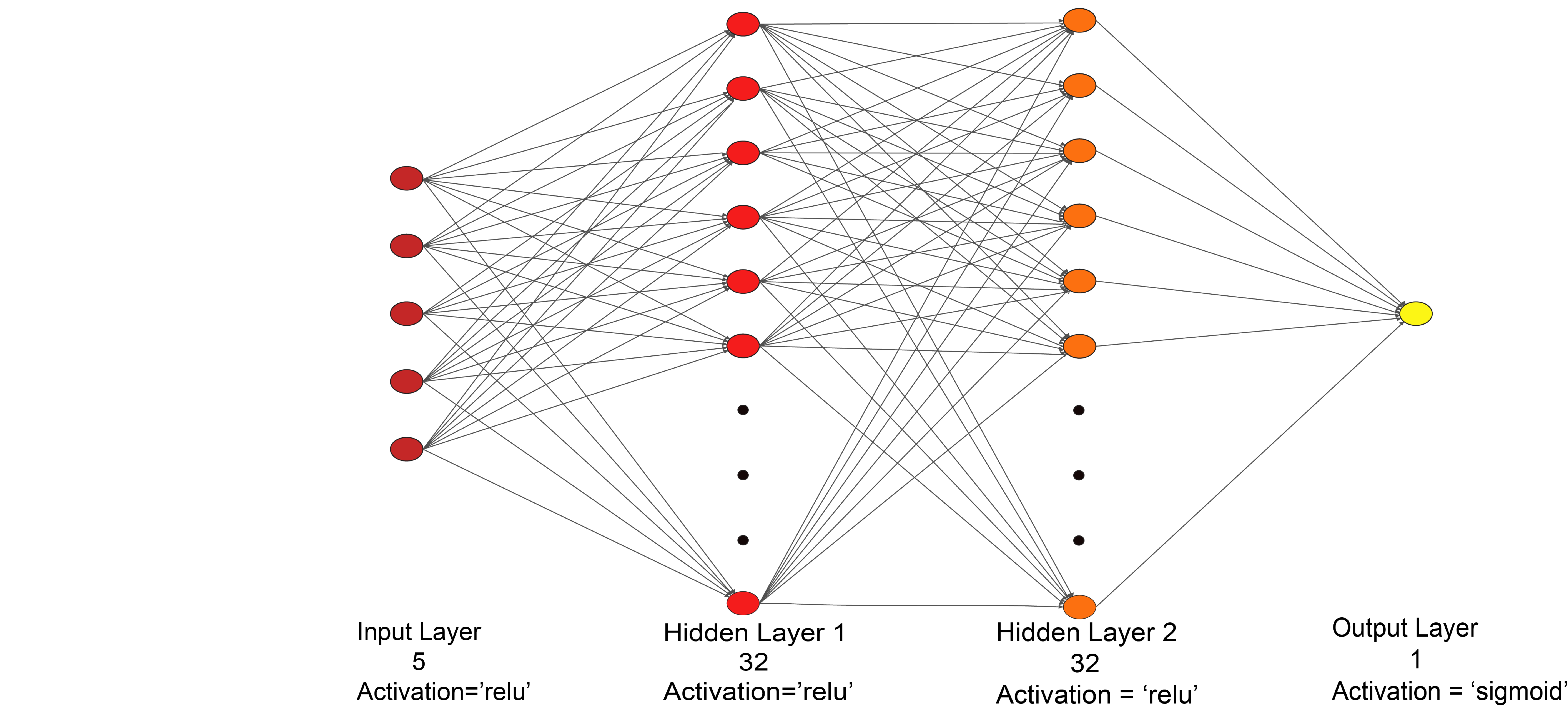}
			\label{neural network}}
		
		%\begin{figure}
		%	\centerline{
		\caption{Proposed model: (a) key processes and (b) neural network architecture.}\label{proposed-model-layers}
	\end{figure*}

Our research  aims at developing an alternative interpretation or translation system for the speech and hearing impaired communication based on lips\hyp state. To this end, we have modelled lips-states streamlining the landmarks, and thus, developed lips-states classifier model using two prominent landmark detectors, and thus, we construct two  models as illustrated in Fig.~\ref{proposed-model-layers}. The first model that we call model-I uses the popular landmark detector \textit{DLIB} to feed data into our neural network based prediction model to predict the lips\hyp states. Similarly, we use the MediaPipe face-mesh landmark detector to develop the second model called model-II with Support Vector Regression (SVR) prediction block. Having similar network architecture, the lips\hyp state  classification accuracy and speed may primarily depend on their underlying detectors. Prior to analysis of these performances in Sec.~\ref{Result}, we now present below the necessary technical details of the proposed models.\looseness -1

	\subsection{Landmark Detection}
	\subsubsection{DLIB.}
	
\begin{figure*}[!thbt]
		\centering 
		\subfloat[]{
			\includegraphics[width=0.3\linewidth,height=3.7cm]{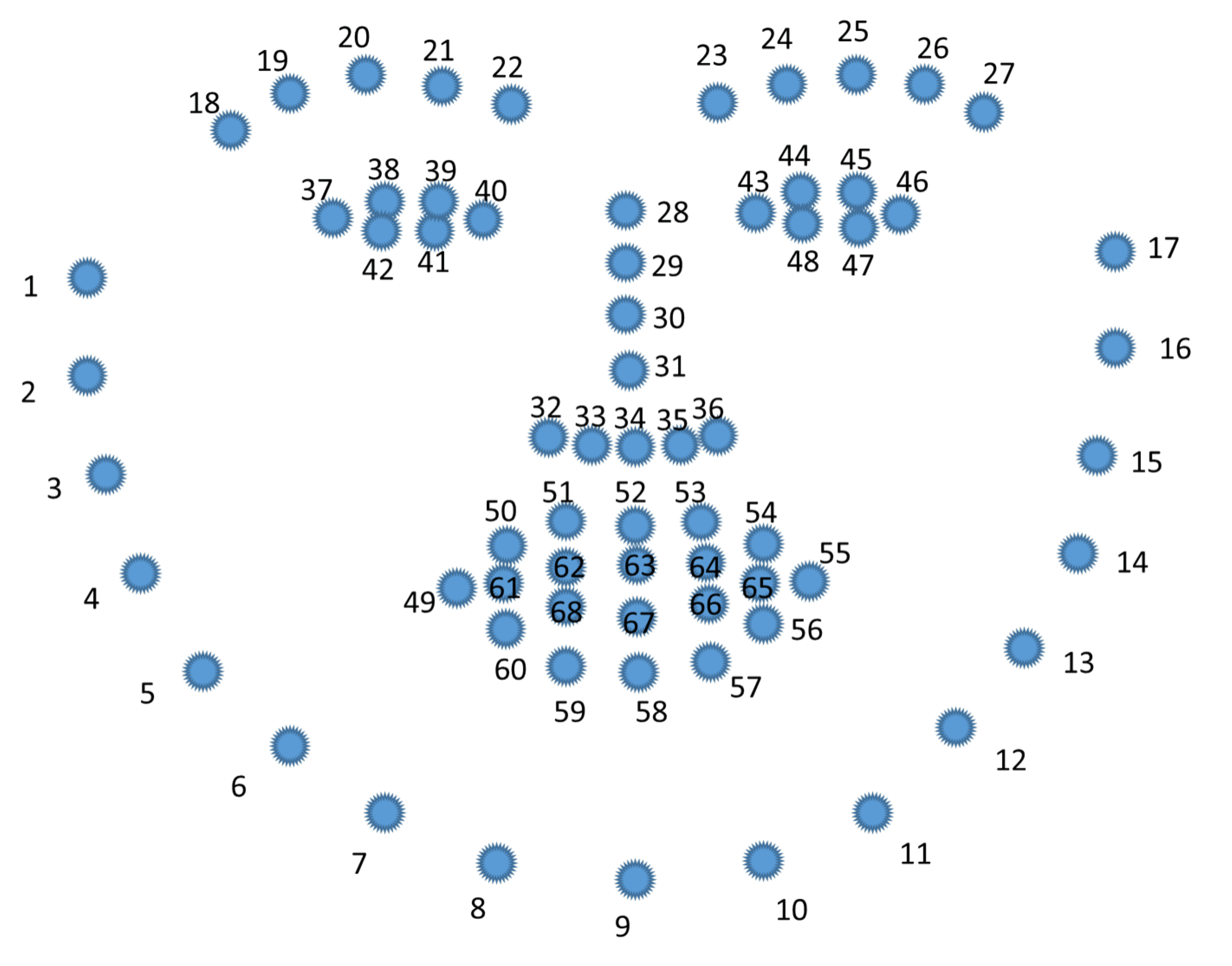}\label{fig:DLIB}
		}\hfill
		%\caption{Facial landmark point in DLIB.}
		%
		%\end{figure}
		%\begin{figure}
		% \centerline{\includegraphics[width=0.9\linewidth, height=5cm]{mediapipe.jpg}}
		\subfloat[]{
			\includegraphics[width=0.3\linewidth,height=3.4cm]{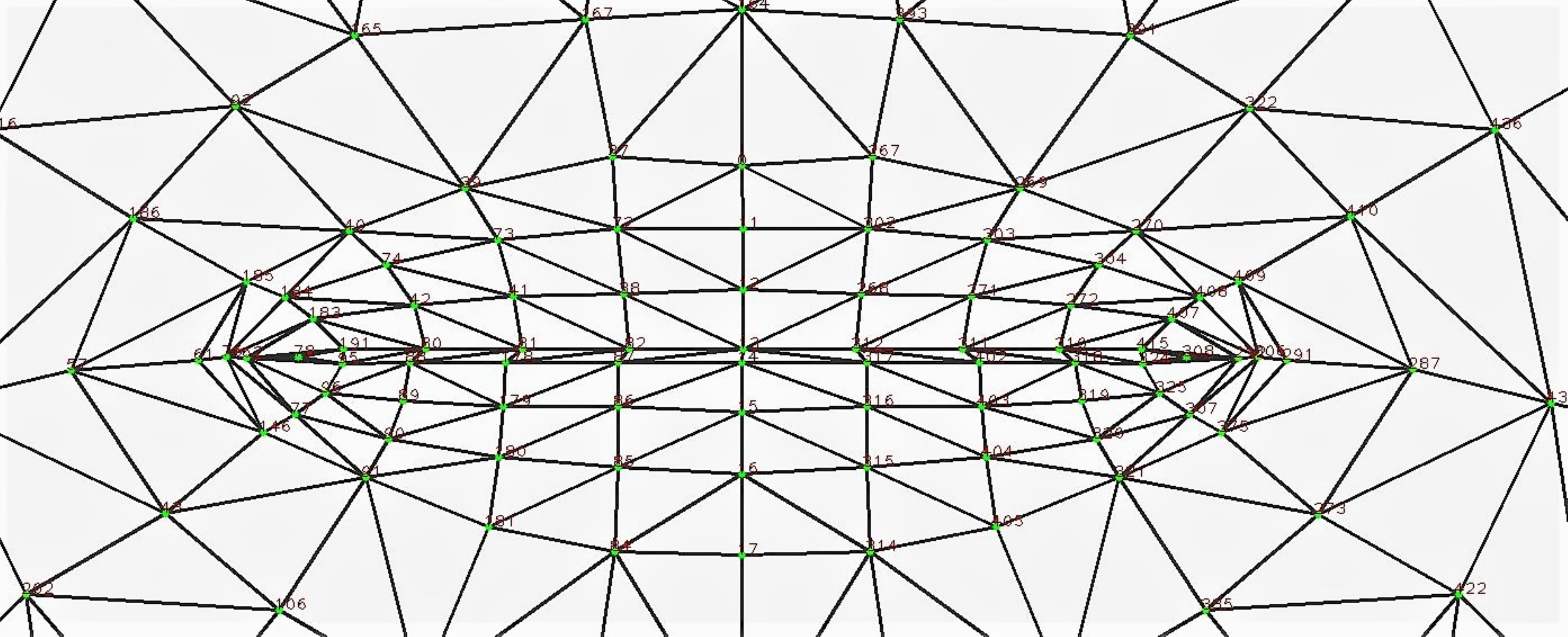}\label{fig:mediapipe}
		}\hfill
		\subfloat[]{
			\includegraphics[width=0.3\linewidth,height=3.4cm]{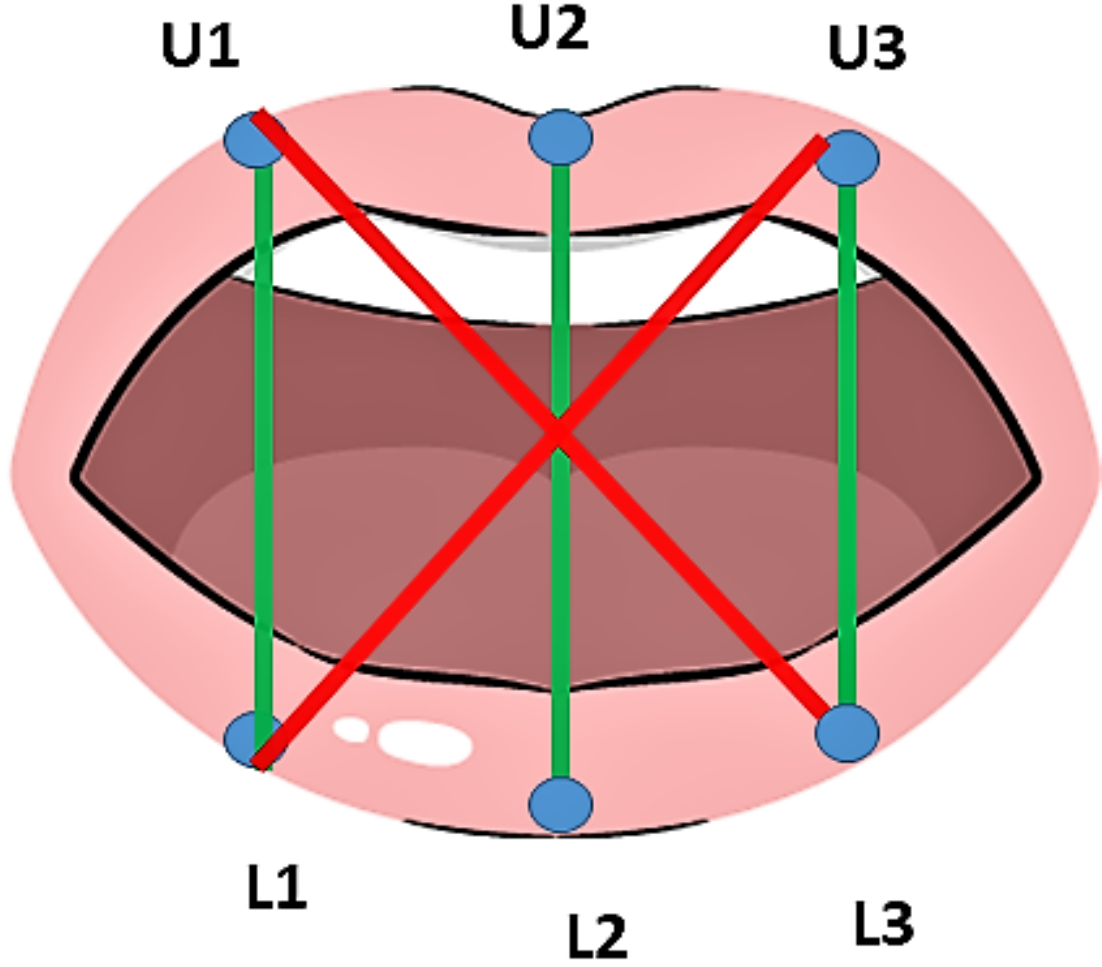}\label{lip landmarks}
		}
		\caption{Facial landmark points: (a)~DLIB, (b)~MediaPipe, and (c)~simplified lips landmarks and distances of our model.}
	\end{figure*}

	For the first model,  pre-trained face feature point detector that comes with the DLIB library is used to obtain 68 Cartesian coordinate points corresponding to a specific area of the face. This 68-points shown in Fig.~\ref{fig:DLIB} comes from the DLIB model which is trained on the iBUG 300-W dataset~\cite{sagonas2016300}. From those landmark point 6 landmark points were selected for lip distance calculation.\looseness -1

	% \begin{figure}[t!]  %subfigg
	%     \centering
	
	%     \subfloat[Detected lip landmark in real time\label{fig:ex1orig}]{%
	%       \includegraphics[width=0.49\linewidth,height=4cm]{7 (3).JPG}}
	%     \hfill
	%     \subfloat[Used lip landmarks and distances\label{fig:ex1gt}]{%
	%         \includegraphics[width=0.49\linewidth,height=4cm]{lip landmarks.png}}
	
	%     \caption{Lip landmarks for real time use. }
	%     \label{lip landmark} 
	% \end{figure}

	\subsubsection{MediaPipe.}
	MediaPipe face mesh is a face geometry detector based on the blaze-face model. It can estimate 468 3D face landmarks as shown in Fig.~\ref{fig:mediapipe}, from both recorded and real-time video streams even on mobile devices. The detector employs machine learning to infer the 3D surface geometry, requiring only a single camera input without the need for a dedicated depth sensor, a feature suitable for our purpose, accurate tracking of lips. Utilising lightweight model architectures together with GPU acceleration throughout the pipeline, the detector also can deliver real-time performance even on weaker processing machines such as \textit{raspberry pi} or \textit{jetson} devices. Out of the 468 landmarks, 6 were used for lips distance calculation.\looseness -1

	\subsection{Distance Calculation}
	After detection of landmarks, five distances were calculated Fig.~\ref{lip landmarks} from the $(x_i,y_i)$ 
	coordinates of the points. Six landmark points were selected for distance calculation shown in Table.~\ref{table landmark}. These distances are computed using Eq.~\eqref{eq1} - \eqref{eq5}.

		\begin{table}[!thb]
		\caption{Landmark used for distance calculation}
		\centering
		\renewcommand{\arraystretch}{1}
		\setlength{\tabcolsep}{8pt}
		\resizebox{0.7\columnwidth}{!}{
			\begin{tabular}{ccc}
				\toprule
				\textbf{Lips landmark point}&\textbf{Our model-I }&\textbf{Our model-II}\\\midrule 
				U1&
				51&
				37\\\midrule
				
				U2&
				52&
				267\\\midrule
				
				U3&
				53&
				0\\\midrule
				
				L1&
				59&
				84\\\midrule
				
				L2 &
				58&
				314\\\midrule
				L3&
				57&
				17\\\bottomrule
				% \vline
			\end{tabular}
		}
		% \caption{Caption}
		\label{table landmark}
	\end{table}

	\vspace{-0.15cm}
	\begin{subequations}
	\begin{equation}\label{eq1}	LD = \alpha \times \sqrt{(U_1[x_1]-L_1[x_2])^2+(U_1[y_1]-L_1[y_2])^2}
	\end{equation}
	\begin{equation}\label{eq2}MD = \beta \times \sqrt{(U_2[x_1]-L_2[x_2])^2+(U_2[y_1]-L_2[y_2])^2}
	\end{equation}
	\begin{equation}\label{eq3}RD = \gamma\times \sqrt{(U_3[x_1]-L3[x_2])^2+(U_3[y_1]-L_3[y_2])^2}
	\end{equation}
	\begin{equation}\label{eq4}D_1= \delta \times \sqrt{(U_1[x_1]-L_3[x_2])^2+(U_1[y_1]-L_3[y_2])^2}
	\end{equation}
	\begin{equation}\label{eq5}D_2= \epsilon \times \sqrt{(U_3[x_1]-L_1[x_2])^2+(U_3[y_1]-L_1[y_2])^2}
	\end{equation}
	\end{subequations}\smallskip

	Here, $\alpha$, $\beta$, $\gamma$, $\delta$, $\epsilon$ are the length control coefficient, which can be tuned for improved observation;
	\begin{math}U_1=\end{math} upper lip left point;
	\begin{math}U_2=\end{math} upper lip middle point;
	\begin{math}U_3=\end{math} upper lip right point;
	\begin{math}L_1=\end{math} lower lip left point;
	\begin{math}L_2=\end{math} lower lip middle point;
	\begin{math}L_3=\end{math} lower lip right point;
	\begin{math}LD =\end{math} lip left distance;
	\begin{math}MD =\end{math} lip middle distance;
	\begin{math}RD =\end{math} lip right distance;
	\begin{math}D1 =\end{math} lip diagonal~1 distance;
	\begin{math}D2 =\end{math} lip diagonal~2 distance.
	These selected landmark points demonstrate the maximum variation upon lips\hyp state . Although the middle points U2 and L2 can be sufficient for our purpose, the other points were considered to tackle the variety of lips shapes during their\hyp states.\looseness -1

	\subsection{Dataset Collection}
	We collected video recording of 15 individuals for training. For each person, video was captured for both closed and open mouth positions. The lips training dataset consists of total 23,412 frames. All of these frames were passed through the preprocessing stage to extract landmarks and calculate  the 5 distances mentioned in distance calculation. Thus a dataset consisting of 5 lips landmark distances and mouth position, either open or close for all frames was created and used to train the data on the train dataset. The validation data consists of 2325 frames of 3 people excluded from the training set.

% \iffalse
% 	\begin{figure}[b]
% 		\centerline{\includegraphics[width=0.7\linewidth, height=4cm]{YOLO.png}}
% 		\caption{Training average loss vs iteration and mAP vs iteration.}
% 		\label{fig2}
% 	\end{figure}
% 	\fi

	\subsection{Lips\hyp State Detection}
	\subsubsection{Proposed model-I using DLIB data.}
	The proposed model-I utilises a typical and very lightweight dense neural net with one dense layer of 32 units and Rectified Linear Unit (RELU) activation. Early stopping was employed for better optimisation of training time and to avoid over-fitting. For compilation of the model, the optimiser was set to Adam and loss was binary cross-entropy. After training and tuning, it was used in real time application.

	Even though the model is very lightweight, the underlying DLIB detector in the data pre\hyp processing stage was adding enough complexity to turn the frames per second (FPS) down. We got about 5 to 9 FPS in real time application, that is usable at best but not robust enough to work in every situation.

	\subsubsection{Proposed model-II using MediaPipe data.}
	Proposed model-II uses the landmark distance values obtained from MediaPipe and plugs it into a Support Vector Classifier (SVC) model. The model was used with the default parameters as it was able to provide sufficiently accurate results.

 \begin{figure}[!t]
		\centerline{
			\includegraphics[width=0.8\textwidth]{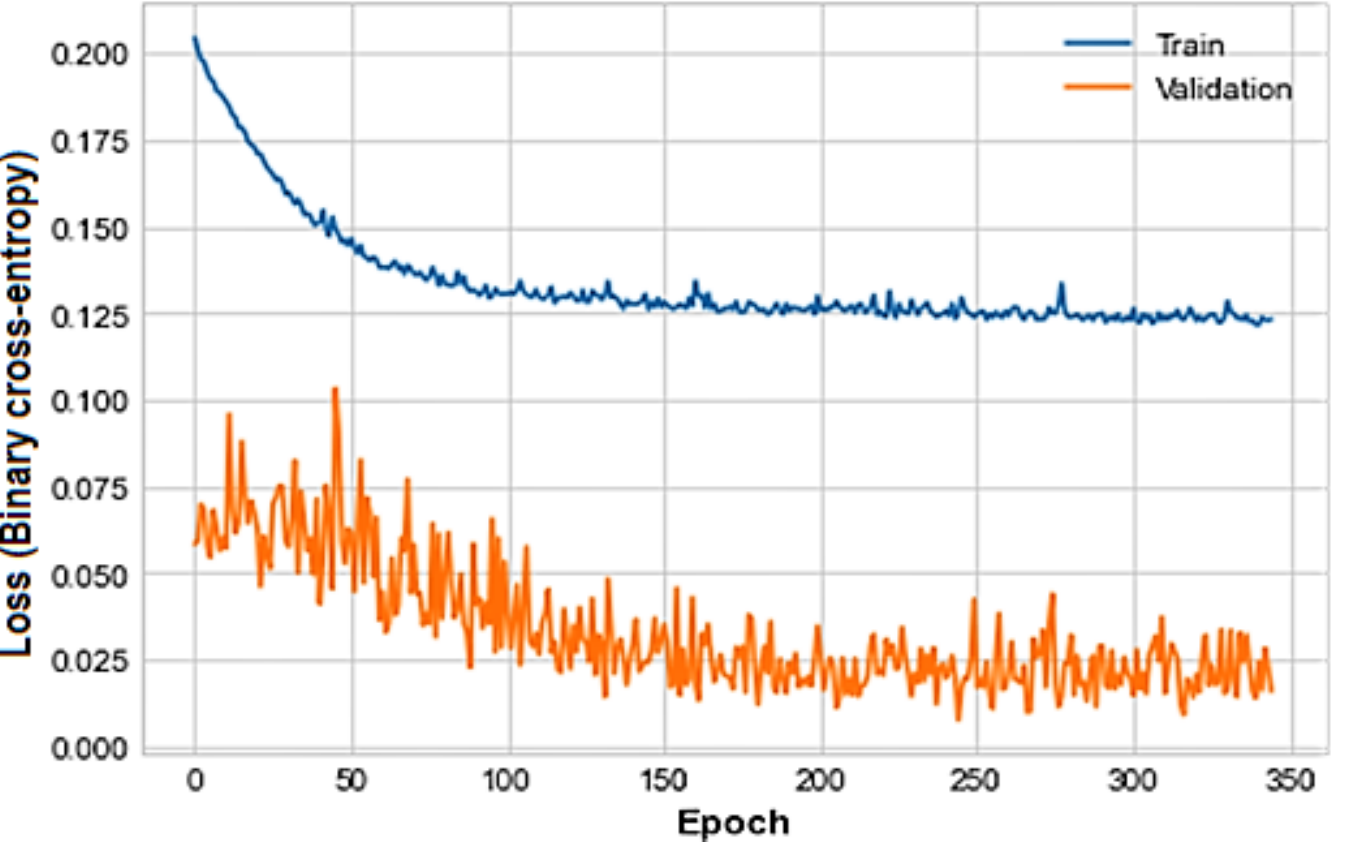}
		}
		\caption{Average loss vs epoch}
		\label{loss}
	\end{figure}

        \begin{figure*}[!t]
		\centering
		\includegraphics[width=0.9\textwidth, height=2in]{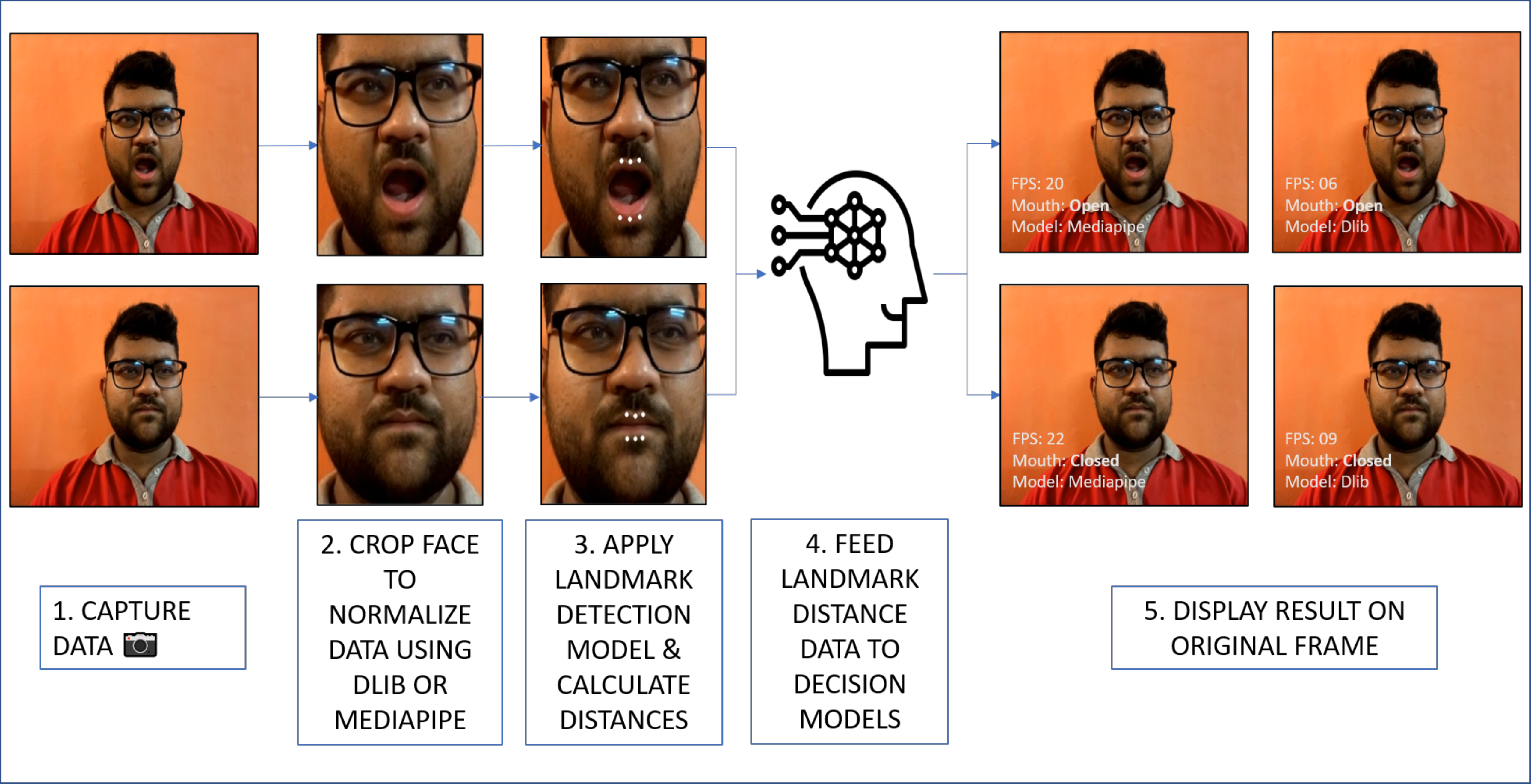}
		\caption{Performance of the proposed model in a realtime application scenario.}
		\label{real time}
	\end{figure*}

	\subsection{Training Performance Evaluation}
	The mouth position detection model was trained for 350 epochs with early stopping employed to stop the model training if the accuracy  improvement  is negligible. It  can  be  seen from  the loss  vs  epoch  curve that the model  is  very  quick  to  get  to  a  lower  loss  value (see Fig.~\ref{loss}).  After  a while  the  improvement as well as  the  learning rate gets saturated.

	\subsection{Time Slots and Commands}
	Each of the detected actions are placed into time slot for converting into commands as shown in Fig.~\ref{translation}. If the time slots are field with N number of actions. The command will be known as N actions per time slot window. The total number of actions per window will increase the maximum number of commands. This is limited by the maximum rate of frame captured and the maximum number of actions taken by the user. Their overall relation can be expressed as follows:
	\begin{equation}T_a=T_s/N_a\end{equation}
	\begin{math}T_s=\end{math} time for \begin{math} N_a;\end{math} \begin{math}T_a=\end{math} time for action; \begin{math}N_a= \end{math} total number of actions per time slot. Since it is binary action, $N_c=2^{Na}$, where \begin{math}N_c=\end{math} total number of commands.
	For example if \begin{math}T_a\end{math} = 0.6 sec, \begin{math}N_a \end{math} = 4, then \begin{math}T_s \end{math} will be \begin{math}0.6 \times 4 = 2.4\end{math}  sec.

	% % \frame{\includegraphics{lipclose.png}}
	% \begin{figure}
	% \includegraphics[width=0.9\linewidth,height=5cm]{lipclose.png}
	% \end{figure}

	\section{Results}
	\label{Result}
	The model was tested on both recorded video footage with video resolution of $1920\times 1080$ at 30 fps as shown in Fig.~\ref{real time} and still images from Flickr-Faces-HQ Dataset (FFHQ)~\cite{karras2019stylebased}.
% 	{\color{red} Explain here what the reader would find in this Fig.~\ref{real time}, and why or how those observations are important } 
{ A model trained with a particular dataset may not give similar results when tested with data from real world. The model is therefore evaluated for the cases unknown to it. 

			\begin{figure}[!tb]  %subfigg
		\centering
		\subfloat[Closed lips\label{fig:ex1orig}]{%
			\includegraphics[width=1\linewidth, height=2.1in]{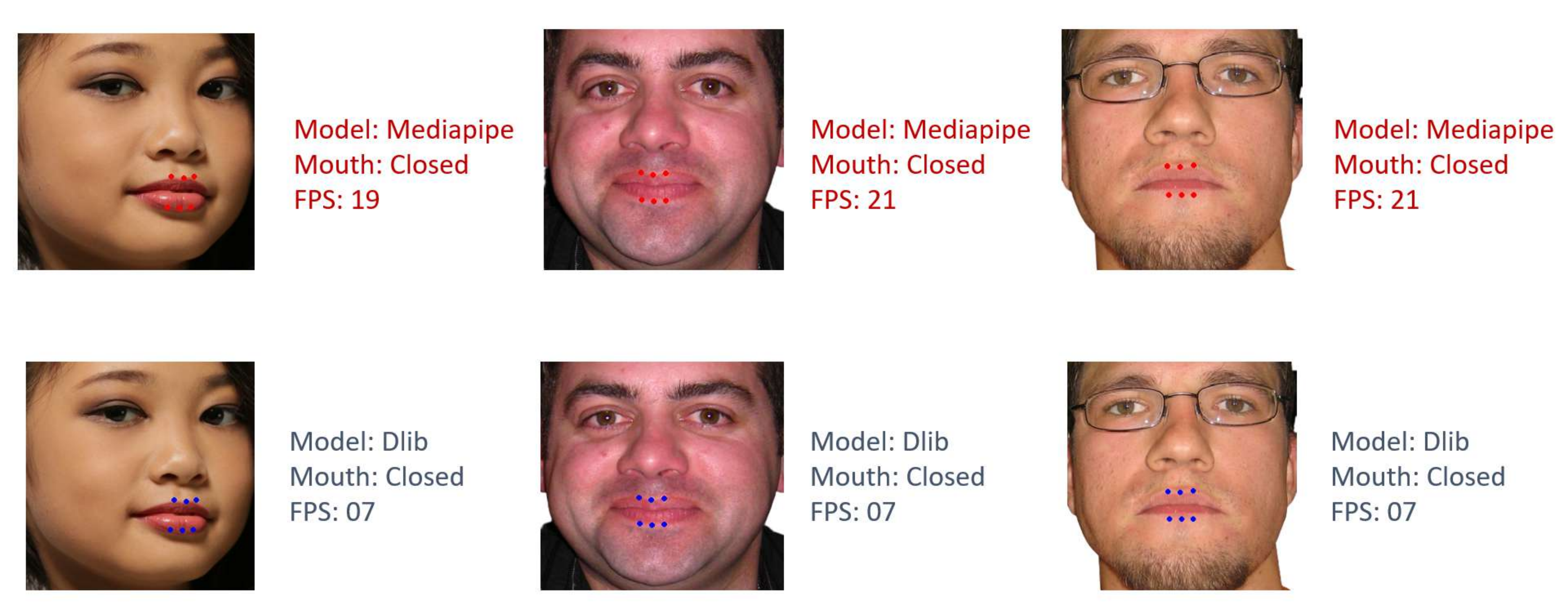}
		}\quad
		%   png}
		%    \hfill
		%\subfloat[Proposed model  1(Closed %lip)\label{fig:ex1gt}]{%
		% \includegraphics[width=0.49\linewidth,height=5cm]%{Dlib_closed.png}}
		%    \hfill
		\subfloat[Open lips\label{fig:ex1resFast}]{%
			\includegraphics[width=1\linewidth, height=2.1in]{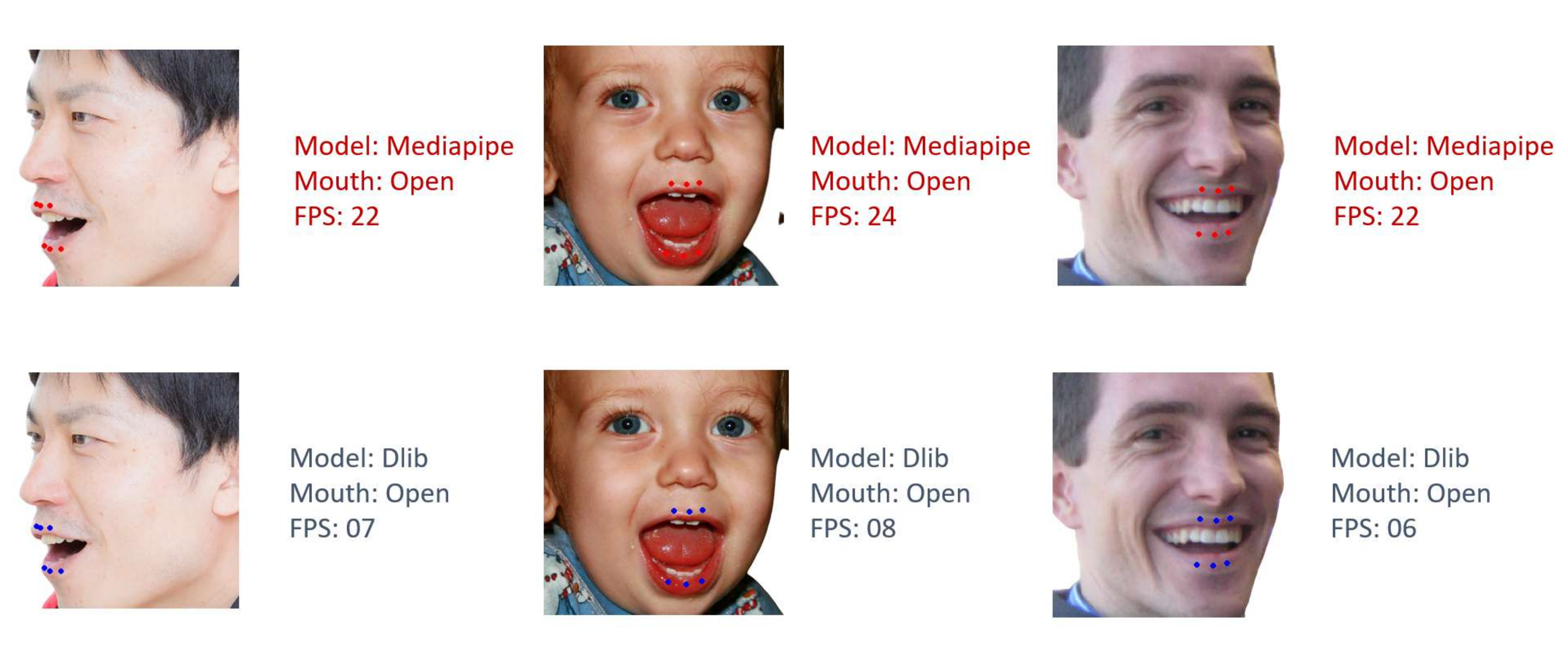}
		}
		%    \hfill
		%\subfloat[Proposed model  2(Closed lip)\label{fig:ex1resEcog}]{%
		% \includegraphics[width=0.49\linewidth,height=5cm]{Mediapipe_closed.png}}
		
		\caption{Examples of lips\hyp state detection using Proposed model-I (\textit{bottom row}) and model-II (\textit{top-row}) for opening and closing lips.}
		\label{real time detection} 
	\end{figure}
	
A step by step working procedure of our models for real-time application scenario is illustrated in Fig~\ref{real time}.
% having an impressive amount of frame rate.
}
	A few cases of visual detection of lips\hyp state ares also illustrated  for the considered datasets in Fig.~\ref{real time detection}. 
% 	{\color{red} Explain here what the reader would find in this figure, and why or how those observations are important}
	{The images in the figure captured varying conditions of face, angle, appearance and lighting 
% 	Some images were sampled based on their rotation and variation in appearance 
	that illustrates the performance of the model for detecting the opening and closing of the lips.
% 	on a well structured dataset. 
% 	The evaluation was made so that comparison can be made in future for similar studies.
	}

	Similarly, the performance of the models while applied on faces turned in different angles have been given in Table.~\ref{tab2}, which shows that the effect of face rotation can be quite significant and similar for both models. The detection accuracy is 100\% from -40 degree to +40 degree angle. But the results are affected when face is rotated beyond that. The accuracy starts declining at +-60 degrees still staying at a quite appreciable 90\%. But if the face is rotated any further, both the models fail to detect the facial landmarks.\looseness-1

	\begin{table}
		\caption{Detection accuracy of proposed model at varying face angles}
		\centering
		\renewcommand{\arraystretch}{0.85}
		\setlength{\tabcolsep}{10pt}
		\resizebox{0.8\columnwidth}{!}{
			\begin{tabular}{ccc}
				% \vline
				\toprule 
				
				\textbf{Angle (degree)}&\textbf{Our model-I\,(\%)} &\textbf{Our model-II\,(\%) }\\
				\midrule
				
				+0&
				100&
				100\\\midrule
				
				+20&
				100&
				100\\\midrule
				
				+40&
				100 &
				100\\\midrule
				
				+60&
				95 &
				95\\\midrule
				
				+80 &
				not detected &
				not detected\\\midrule
				-0&
				100&
				100\\\midrule
				
				-20&
				100&
				100\\\midrule
				
				-40&
				100 &
				100\\\midrule
				
				-60&
				95 &
				95\\\midrule
				
				-80 &
				not detected &
				not detected\\
				\bottomrule
				% \vline
			\end{tabular}
		}
		% \caption{Caption}
		\label{tab2}
	\end{table}

	\begin{table}
		\caption{Detection accuracy at varying lips\hyp state detection rate}
		\centering
		\renewcommand{\arraystretch}{0.85}
		\setlength{\tabcolsep}{8pt}
		\resizebox{0.8\columnwidth}{!}{
			\begin{tabular}{ccc}
				% \vline
				\toprule 
				\textbf{\makecell[c]{Movements \\per sec}}& \textbf{Our model-I\,(\%)} & \textbf{Our model-II\,(\%)}\\
				\midrule
				
				1&
				100&
				100\\\midrule
				
				2&
				100&
				100\\\midrule
				
				3&
				95 &
				100\\\midrule
				
				4&
				85 &
				95\\\midrule
				
				5 &
				75 &
				90\\\midrule
				6&
				60&
				80\\
				\bottomrule
				% \vline
			\end{tabular}
		}
		% \caption{Caption}
		\label{tab3}
	\end{table}
	
	\begin{table}
		\caption{Overall performance of lips\hyp state detection}
		\centering
		\renewcommand{\arraystretch}{1}
		\setlength{\tabcolsep}{10pt}
		\resizebox{0.8\columnwidth}{!}{
			\begin{tabular}{lcc}
				% \vline
				\toprule
				
				\multicolumn{1}{c}{\textbf{Parameters}}&\textbf{Our model-I}&\textbf{Our model-II}\\
				\midrule
				Best Validation Accuracy&
				95\%&
				94\%\\\midrule
				
				Best Training Loss&
				0.122\%&
				0.130\%
				\\\midrule
				
				Best Training Accuracy&
				95.47\%&
				94.808\%\\\midrule
				
				Average Accuracy&
				95.25\%&
				94.40\%\\\midrule
				
				Average FPS&
				6&
				20\\
				\bottomrule
				% \vline
			\end{tabular}
		}
		% \caption{Caption}
		\label{tab4}
	\end{table}

	\begin{table}
		\caption{Sample of training data}
		\centering
		\renewcommand{\arraystretch}{1}
		\setlength{\tabcolsep}{10pt}
		\resizebox{0.9\columnwidth}{!}{
			\begin{tabular}{cccccc}
				\toprule
				\textbf{Left}& \textbf{Middle }& \textbf{Right}&\textbf{Diagonal-1}&\textbf{Diagonal-2}&\textbf{Output}\\\midrule
				30.06659&29.123&28.01785&31.38471&33.10589&1\\\midrule
				30.06659&30.01666&29.01724&32.31099&33.54102&1\\\midrule
				34.05877&34.0147&33.01515&36.05551&37.16181&1\\\midrule
				17.02939&17.02939&17.02939&21.40093&21.40093&0
				\\\midrule
				18.02776&17.03876&17.02939&21.63331&21.40093&0\\\midrule
				18.02776&17.01345&16.03122&21.63331&21.40093&0\\
				\bottomrule
			\end{tabular}
		}
		% \caption{Caption}
		\label{training data}
	\end{table}

	The models' accuracy at different speeds of changing the lips-states is also evaluated and given in the Table.~\ref{tab3}. We observed that the accuracy decreases for both models with the increase in the lips\hyp state per second. 
	{For instance, as in Table~\ref{tab3}, both of the models can detect the lips-states, while the lips open/close at a rate of 2 per second. For a different case, when the rate increases to six per second, nearly 40\% of the state-changes are not detected.}
	However, compared to model-I, the model-II is found more accurate for the states being changed at a higher rate.\looseness -1
	
	The overall training and test accuracy of the models is given in Table.~\ref{tab4}, where we observed that model-I is slightly more accurate, while model-II is much faster, almost 3 times than model-I.
	So, proposed model-II can be a better choice for fast moving scenarios. In contrast, model-I is more suited for situations where accuracy is the main priority.
	{In other words, Table~\ref{tab4} indicates that with only 6 frames per second, model-I has an average accuracy of 95.25\%, and with 20 frames per second, model-II has an average accuracy of  94.40\%.}
	
	Additionally, some values of the dataset are given in  Table.~\ref{training data} that suggest that the higher distance values correspond to mouth open position whereas lower distance values normally indicate closed mouth position. 
% 		{\color{red} Can you please further explain this, taking an instance from the table}
{For example, in Table~\ref{training data}, we see that the maximum distance between the landmarks are for \textit{diagonals} and the values are comparatively higher in case of open lips than that of the closed lips. The shortest distances, in contrast, are between the landmarks pair of leftmost and rightmost sides, and they are relatively lower for the case of closed lips.}\looseness -1
	
	Finally, the lips\hyp state detection for varying conditions are illustrated in Fig.~\ref{real time detection}. 
	The trends in accurately detecting the lips-states also holds for the other test images. 
	Capability of accurately detecting the lips opening and closing at different angles and lighting conditions means that the proposed model have the potential for a lips\hyp state based electronic translator of a nonverbal communication system.
% 	instances represented there also hold capability of our model to accurately detect lips movement and determine mouth position even in varying lighting conditions.
	\looseness-1

	\section{Conclusion}
	In support of developing an alternative interpretation/translation system for non-verbal communication system, we have introduced two new models for lips\hyp state detection. Building upon two popular facial landmark detectors, DLIB and MediaPipe, the proposed models have been investigated for the classification of opening and closing of lips for the standard datasets. Being faster, computationally efficient (in terms of FPS) and reasonably accurate, the proposed model-II with MediaPipe can be a promising candidate for the envisaged translation system. Our research continues to further develop the detection accuracy of the proposed models, particularly for extreme facial rotations and very poor lighting conditions.  
	\\
	\\
	\\
	
	\balance
	\bibliographystyle{splncs04}
	\bibliography{212}

\begin{thebibliography}{10}
\providecommand{\url}[1]{\texttt{#1}}
\providecommand{\urlprefix}{URL }
\providecommand{\doi}[1]{https://doi.org/#1}

\bibitem{amornpan2019face}
Amornpan, P., Praisan, P.: Face recognition using transferred deep learning for
  feature extraction. In: 2019 Joint International Conference on Digital Arts,
  Media and Technology with ECTI Northern Section Conference on Electrical,
  Electronics, Computer and Telecommunications Engineering (2019)

\bibitem{bouvier}
Bouvier, C., Benoit, A., Caplier, A., Coulon, P.Y.: {Open or Closed Mouth State
  Detection: Static Supervised Classification Based on Log-polar Signature}.
  In: {ACIVS 2008 - International Conference on Advanced Concepts for
  Intelligent Vision Systems}. vol. Volume 5259/2008, pp. 1093--1102. {Springer
  Berlin / Heidelberg}, Juan-Les-Pins, France (Oct 2008),
  \url{https://hal.archives-ouvertes.fr/hal-00372148}

\bibitem{karras2019stylebased}
Karras, T., Laine, S., Aila, T.: A style-based generator architecture for
  generative adversarial networks (2019)

\bibitem{dlib09}
King, D.E.: Dlib-ml: A machine learning toolkit. Journal of Machine Learning
  Research  \textbf{10},  1755--1758 (2009)

\bibitem{krause2020automatic}
Krause, P.A., Kay, C.A., Kawamoto, A.H.: Automatic motion tracking of lips
  using digital video and openface 2.0. Laboratory Phonology: Journal of the
  Association for Laboratory Phonology  \textbf{11}(1) (2020)

\bibitem{Lu}
Lu, Y., Li, H.: Automatic lip-reading system based on deep convolutional neural
  network and attention-based long short-term memory. Applied Sciences
  \textbf{9}(8), ~1599 (Apr 2019). \doi{10.3390/app9081599},
  \url{http://dx.doi.org/10.3390/app9081599}

\bibitem{lugaresi2019mediapipe}
Lugaresi, C., Tang, J., Nash, H., McClanahan, C., Uboweja, E., Hays, M., Zhang,
  F., Chang, C.L., Yong, M.G., Lee, J., et~al.: Mediapipe: A framework for
  building perception pipelines. arXiv preprint arXiv:1906.08172  (2019)

\bibitem{sagonas2016300}
Sagonas, C., Antonakos, E., Tzimiropoulos, G., Zafeiriou, S., Pantic, M.: 300
  faces in-the-wild challenge: Database and results. Image and vision computing
   \textbf{47},  3--18 (2016)

\bibitem{sharma2016farec}
Sharma, S., Shanmugasundaram, K., Ramasamy, S.K.: Farec—cnn based efficient
  face recognition technique using dlib. In: 2016 International Conference on
  Advanced Communication Control and Computing Technologies (ICACCCT). pp.
  192--195. IEEE (2016)

\bibitem{singh2016improved}
Singh, B., Sahoo, S., Kumar, V., Issac, A., Dutta, M.K.: Improved lip contour
  extraction using k-means clustering and ellipse fitting. In: 2016 2nd
  International Conference on Communication Control and Intelligent Systems
  (CCIS). pp. 99--103. IEEE (2016)

\bibitem{wenjuan2010real}
WenJuan, Y., YaLing, L., MingHui, D.: A real-time lip localization and tacking
  for lip reading. In: 2010 3rd International Conference on Advanced Computer
  Theory and Engineering (ICACTE). vol.~6, pp. V6--363. IEEE (2010)

\bibitem{xu2020anchorface}
Xu, Z., Li, B., Geng, M., Yuan, Y., Yu, G.: Anchorface: An anchor-based facial
  landmark detector across large poses. arXiv preprint arXiv:2007.03221  (2020)

\end{thebibliography}

\end{document}